\documentclass[journal]{IEEEtran}
\UseRawInputEncoding 
\usepackage{graphicx}
\usepackage{amsmath,amssymb,amsfonts,dsfont}
\usepackage{stmaryrd}
\usepackage{url}
\usepackage{dsfont}
\usepackage{float}
\usepackage{hyperref}
\usepackage{multirow}
\usepackage{adjustbox}
\usepackage{graphicx}
\usepackage{siunitx}
\usepackage{hhline}
\usepackage[dvipsnames,table,xcdraw]{xcolor}
\usepackage[caption=false]{subfig}
\usepackage{caption}
\ifCLASSINFOpdf
\else
\fi

\begin{document}
\title{RobustSleepNet: Transfer learning for automated sleep staging at scale}

\author{Antoine~Guillot and~Valentin~Thorey% <-this % stops a space
\thanks{A. Guillot and V. Thorey are with the Algorithm Team, Dreem, Paris

}}

\maketitle

\begin{abstract}
Sleep disorder diagnosis relies on the analysis of polysomnography (PSG) records. As a preliminary step of this examination, sleep stages are systematically determined. In practice, sleep stage classification relies on the visual inspection of 30-second epochs of polysomnography signals. Numerous automatic approaches have been developed to replace this tedious and expensive task. Although these methods demonstrated better performance than human sleep experts on specific datasets, they remain largely unused in sleep clinics. The main reason is that each sleep clinic uses a specific PSG montage that most automatic approaches cannot handle out-of-the-box. Moreover, even when the PSG montage is compatible, publications have shown that automatic approaches perform poorly on unseen data with different demographics.
To address these issues, we introduce RobustSleepNet, a deep learning model for automatic sleep stage classification able to handle arbitrary PSG montages. We trained and evaluated this model in a leave-one-out-dataset fashion on a large corpus of 8 heterogeneous sleep staging datasets to make it robust to demographic changes.
When evaluated on an unseen dataset, RobustSleepNet reaches 97\% of the F1 of a model explicitly trained on this dataset. Hence, RobustSleepNet unlocks the possibility to perform high-quality out-of-the-box automatic sleep staging with any clinical setup.
We further show that finetuning RobustSleepNet, using a part of the unseen dataset, increases the F1 by 2\% when compared to a model trained specifically for this dataset. Therefore, finetuning might be used to reach a state-of-the-art level of performance on a specific population.
\end{abstract}


% \fi
\IEEEpeerreviewmaketitle

\section{Introduction}\label{introduction}
During sleep, the body alternates through different physiological states called sleep stages. The AASM (American Academy of Sleep Medicine) guidelines define five different sleep stages: wake, NREM1, NREM2, NREM3 (or deep sleep), and REM (rapid eye movement) \cite{Iber2007TheSpecifications}. Sleep stage classification is a primary tool used systematically in the diagnosis of sleep disorders such as narcolepsy \cite{Stephansen2018} or sleep apnea \cite{Thorey_2019}.
Sleep stages are determined with the use of a polysomnography (PSG) device. It can monitor several physiological signals throughout the night, but the ones typically recorded are brain activity  (electroencephalography [EEG]), muscle activity (electromyography [EMG]), eye movements (electrooculography [EOG]), cardiac activity (electrocardiography [ECG]), and breathing.
In a clinical setup, sleep stage classification is generally done manually by a trained sleep expert who visually analyses the recorded signals. The analysis is conducted on 30-seconds epochs, where each epoch is associated with one of the five sleep stages according to the standards defined by the AASM. The manual classification is quite expensive, as it takes up to an hour to score a whole recording (which is around 1000 epochs or 8-hours long).
The sleep scoring guidelines do not fully reflect the variability of patterns that exist among subjects. As a consequence, different scorers can interpret the same epoch differently. An analysis of nights scored by more than 2,500 scorers shows an average inter-rater agreement of 82.6\%  across sleep stages  \cite{Rosenberg2014TheEvents}, the inter-rater agreement strongly depends on the dataset demographics and on the sleep center where the study is recorded. As a rule of thumb, a dataset with older subjects or subjects with sleep disorders presents lower inter-rater agreements \cite{guillot2019dreem} than a dataset composed of healthy subjects \cite{10.1093/sleep/zsaa097}.

The high cost and relatively important bias in manual sleep staging have raised an interest in developing automatic sleep staging systems. Recently, deep learning has successfully reached human-performances in automatic sleep staging, in a supervised learning context \cite{supratak2017deepsleepnet} \cite{SeqSleepNet} \cite{guillot2019dreem} \cite{Stephansen2018} \cite{perslev2019utime} \cite{phan2021xsleepnet}; despite this, automated sleep staging methods remain unused in most sleep clinics. We identified two important factors to explain why sleep clinics still rely on manual sleep staging. We will refer to these using the terms "input-shape incompatibility" and "accuracy-drop issue."

% Problematic
The input-shape incompatibility issue is practical. Each sleep clinic has a specific PSG montage that can differ according to the device used, the particularity of the targeted population, or specific requirements from a study. Automatic approaches can handle only a specific input shape — they are trained on a specific dataset with a specific montage — a different PSG montage will likely generate an incompatible input shape and thus the impossibility to compute an output for the model. 

Conversely, the accuracy-drop issue occurs when a PSG montage is compatible between two datasets. In that case, a model trained on the first dataset is able to compute an output on the second dataset but is likely to perform poorly. This drop in performance is thought to be due to a difference in demographic distribution between the two datasets; however, another causative factor might be channel mismatch. Indeed, the recorded signals are likely to differ (e.g., different EEG positioning, different sensor technology.) In \cite{guillot2019dreem}, the cohen's kappa ($\kappa$) of a model trained on healthy subjects drops from 84.6 to 45.7\% when evaluated on subjects with sleep apnea; the drop was only 82.3 to 77.6 when the model was trained on people with apnea and evaluated on healthy subjects. Similarly, in \cite{Optimal_transport_chambon}, the balanced accuracy of a model trained on a dataset with older subjects drops by 10 points when evaluated on a dataset of young and healthy subjects. In \cite{MultipleChannelTrasnferLearning} models trained on the EEG and EOG derivations of the MASS dataset \cite{MASS} and evaluated on the Sleep-EDF datasets \cite{SleepEDF} experience macro-F1 drop of 11 points compared to their supervised learning counterpart. This drop is likely due to the EEG derivations being different in the two datasets — the drop increases to more than 20 points when the model is trained on EEG and evaluated on the EOG channels.

%  SOTA
To our knowledge, the input-shape incompatibility has never been addressed in the literature; however, several approaches have been used to try to tackle the accuracy-drop issue. In \cite{SingleChannelMismatch} and \cite{MultipleChannelTrasnferLearning}, the models were pre-trained on the Montreal archive of sleep studies (MASS) dataset and finetuned on the Sleep-EDF dataset; finetuning increases the performance reached in a supervised learning setting by 3 to 5 points.
Similarly, in \cite{KLPersonalization}, the model was pre-trained on MASS, finetuned on the first night from a subject from Sleep-EDF and evaluated on the second night; finetuning increased performances by 6 points, and the finetuned model was only one point below the supervised learning model. The authors used strong regularization to avoid overfitting. In \cite{Optimal_transport_chambon}, the authors aligned the observations from the source and target dataset to reduce the distribution shift between them. This method does not require labels from the target dataset, but the balanced accuracy is 10 points below the supervised baseline.
However, all these methods use part of the target dataset for training, which does not remove the high fixed cost of retraining and adapting the model for a specific target dataset. Hence, they remain unaffordable for most sleep clinics.

%  Our methods
In this study, we aimed to design an approach that can be used out of the box on unseen data in sleep clinics.
We first tackle the input-shape incompatibility by introducing the RobustSleepNet, a deep learning model that can handle any PSG montage, thus any input shape. This input-invariance property allows us to tackle the second issue: the accuracy-drop on unseen data. We greedily address this issue by training our model on a large corpus of eight datasets with substantial input sensors and demographic variability to increase the generalization capacity of the model.

In the following section, we first present the general architecture and principles of RobustSleepNet and the height datasets used in this study. We evaluate our approach in various setups and compare our results to available literature results. Specifically, we evaluate the performance of our approach in the case of direct transfer to unseen data. We complete this evaluation with several experiments designed to evaluate the impact of some important experimental parameters.

\newcolumntype{L}{>{\centering\arraybackslash}m{3cm}}
\newcolumntype{X}{>{\centering\arraybackslash}m{4cm}}
\begin{table*}[t]
\centering
\scalebox{0.95}{%
\begin{tabular}{l|lLllllXl} 

 \hline
Dataset  & Nights & Channels used & Male (\%) & Age & BMI & AHI & Other disorders & TST (min) \\
 \hline
DOD-O & 55 & 8xEEG, 2xEOG, EMG & 63.6 &$45.6 \pm 16.5$  & $29.6 \pm 6.4$  & $18.5 \pm 16.2$ & - & 387 \\
DOD-H & 25 & 12xEEG, 2xEOG, EMG & 76.0 & $35.3 \pm 7.5$ & $23.8 \pm 3.4 $ & $\leq 5$ & - & 427 \\
MASS  & 115 & 3xEEG, 2xEOG, EMG & 53.9 &$52.2 \pm 17.8$ & - & $\leq 20$ & - & 426 \\
Sleep-EDF & 197 & 2xEEG, EOG & 56 & $54.7 \pm 22.6 $  & - & - & INS(44), TEM(22)  & 426 \\

MrOS & 586 & 5xEEG, EOG, 2xEMG  & 100 & $81.0 \pm 4.4$& $26.9	\pm 3.9$ & $17.1 \pm 15.6$  & PD(16), AD(15), Narcolepsy(1), RLS(25), PLM(11) & 341 \\
MESA & 2053 & 3xEEG, 2xEOG, EMG & 53.5 & $69.6 \pm 9.2$ & - &  $16.1 \pm 13.7$ & - & 359\\
SHHS & 2651 & 2xEEG, 2xEOG, EMG & 45.6 & $68.3 \pm 10.5$ & $28.3 \pm 5.1$  & $14.6 \pm 12.0$ & NARC(9), INS(46), PLM(56)  & 371 \\
CAP & 106 & 2xEEG, EOG & 61& $45.2 \pm 19.6$ & - & - & NFLE(40), PLM(10), RBD(22), BRUX(2), NARC(5), INS(9)  & 410 \\

 \hline
 
\end{tabular}%
}
\caption{Datasets demographics. All values are averaged across subjects. When AHI is reported, we report the mean AHI and its deviation. If AHI is used as an exclusion criterion, we report this criterion. Otherwise, the cell is left blank. We report sleep disorder and the number of subjects said to have been diagnosed for this pathology in parenthesis. Hence all pathologies may not be reported. \\ \small{ AD: Alzheimer's diseases. BRUX: bruxism. INS: insomnia. NARC: Narcolepsy. NFLE: Nocturnal frontal lobe epilepsy. PLM: Periodic leg movements. PD: Parkinson's disease. RBD: REM behavior disorder. RLS: Restless legs syndrome. TEM: Temazepam medications.}}

\label{tab:dataset_desctiption}
\vspace{-1 em}
\end{table*}
\section{Methods}

\subsection{Problem definition}
\label{sec:problem_def}

We represented the signals acquired by the PSG during a night as a sequence of 30-second epochs in $\mathbb{R}^{C \times L}$. $C$ is the number of channels of the PSG, and $L = 30 \times f_s$ the number of points in the epoch. $f_s$ is the sampling frequency of the PSG signal. We assumed that $f_s$ was the same for every signal because our preprocessing included resampling each signal to the same frequency. We labeled each epoch with one of the five sleep stages, which were subsequently converted using one-hot encoding in $(0,1)^5$.

We defined a dataset $\mathcal{D}$ as a set of $N$ scored PSG nights from different subjects recorded in the same conditions. Notably, the number of channels $C_\mathcal{D}$ and the segment length $L_\mathcal{D}$ was the same for every record within the dataset $\mathcal{D}$.

\begin{figure}[t]
\vspace{-1em}
    \begin{center}
    \includegraphics[width=\linewidth]{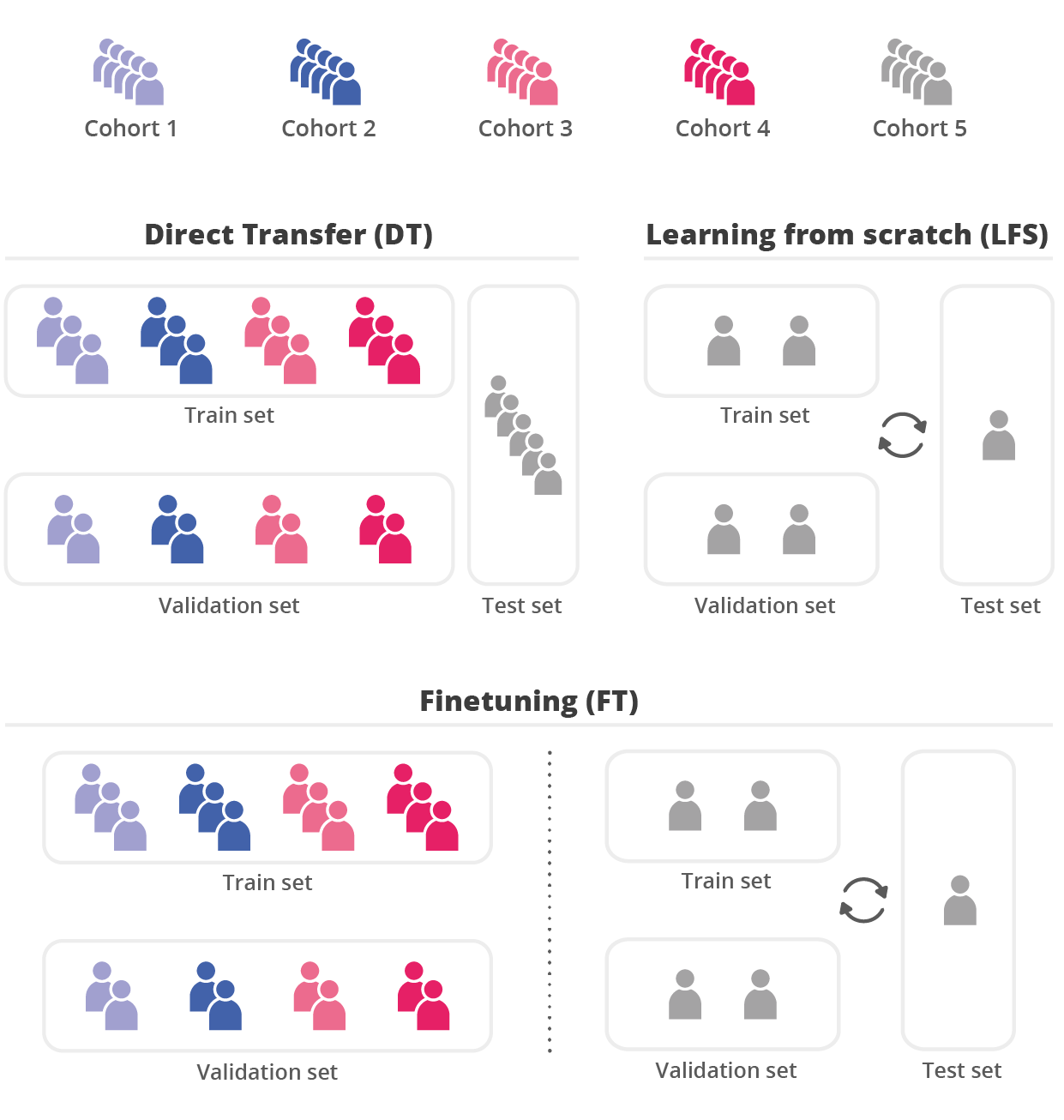} 
    \caption{Illustration of the three learning settings while working with five datasets and five subjects per dataset. The rotating arrow symbolizes the need to use cross-validation to evaluate the performance of the whole test set (FT and LFS).}
    \label{figure:learning_type}
    \end{center}
    \vspace{-2em}
\end{figure}

The problem was predicting the sleep stages of every night from a dataset $\mathcal{D}_{target}$  using a trainable deep learning model and a set of several source datasets $D_{sources} = \{\mathcal{D}_{1}, \mathcal{D}_{2}, ..\}$. To do so, we benchmarked our model using three learning settings (Figure \ref{figure:learning_type}): 
\begin{enumerate}
    \item Direct Transfer (DT): we trained the model using the $D_{sources}$ and evaluated on $\mathcal{D}_{target}$. This is the most interesting approach since we did not use any data from $\mathcal{D}_{target}$ during training.
    \item Learning from scratch (LFS): we trained and evaluated the model on the dataset $\mathcal{D}_{target}$ in a k-folds fashion. Most automatic sleep stage classification systems use this approach \cite{SeqSleepNet} \cite{supratak2017deepsleepnet} \cite{guillot2019dreem}. We split the data subject-wise to avoid data leakage: when several nights were available for one subject, they were assigned to the same fold.
    \item Finetuning (FT): this is a mix of both previous approaches. We first trained the model using the $D_{sources}$ as in the DT approach. We then trained and evaluated the model on the $\mathcal{D}_{target}$ dataset in a k-folds fashion similar to the LFS setting.
\end{enumerate}
  
\subsection{Evaluation}
\label{part_2_evaluation}
Once the deep learning model had predicted the sleep stages for every epoch of every night from the dataset $\mathcal{D}_{target}$ we could compute a performance metric. To compare with approaches from the literature, we used the Macro-F1 score over all the epochs from a dataset.
$$F1 = \frac{1}{5} \sum_{1 \leq i \leq 5} \frac{precision_i . recall_i}{precision_i + recall_i}$$ where $precision_i$ and $recall_i$ respectively denoted the precision and recall for the i-th sleep stage.
We will refer to the Macro-F1 in this paper using $F1$.

\subsection{Measure of the dataset generalization and easiness}
\label{section:easiness_and_gen}
To characterize the individual contribution of each dataset used in this study during DT, we defined two metrics called generalization and easiness.

Given $N_{datasets}$ datasets, $F1(\mathcal{D}_i,\mathcal{D}_j)$ denotes the F1 of a model trained on the i-th dataset, and evaluated on the j-th dataset using the DT setting. We introduced $F1_{rel}(\mathcal{D}_i,\mathcal{D}_j)$ the relative F1 of a model trained on $\mathcal{D}_i$ and evaluated on $\mathcal{D}_j$ with respect to the LFS performance of this model $F1_{\mathcal{D}_j}$.

$$F1_{rel}(\mathcal{D}_i,\mathcal{D}_j) = \frac{F1(\mathcal{D}_i,\mathcal{D}_j)}{F1_{\mathcal{D}_j}}$$
$$ \text{generalisation}(\mathcal{D}_k) = \frac{1}{N_{datasets} - 1}\sum_{i \neq k} F1_{rel}(\mathcal{D}_k,\mathcal{D}_i) $$

$$ \text{easiness}(\mathcal{D}_k) = \frac{1}{N_{datasets} - 1}\sum_{i \neq k} F1_{rel}(\mathcal{D}_i,\mathcal{D}_k)  $$

The generalization of $\mathcal{D}$ quantifies how well a model trained on $\mathcal{D}$ performs on each dataset from $D_{sources}$.

The easiness of $\mathcal{D}$ quantifies how well the models trained on each $D_{sources}$ perform on this $\mathcal{D}$. 

The use of $F1_{rel}$ tackles the fact that some datasets inherently present lower or higher absolute performance in every setting (DT, FT, and LFS). Normalizing F1 ensures a fair comparison in that case.

\subsection{RobustSleepNet overview}

In this publication, we formulated the problem of sleep stage classification as a sequence-to-sequence problem. Indeed, instead of considering one epoch at a time, we used a temporal context of $T$ successive epochs and tried to predict the associated $T$ sleep stages. Hence, we represented the input signal domain as $\mathbf{Z} \in \mathbb{R}^{ C \times L \times T}$. The output  $\hat{\pi} \in [0, 1]^{5 \times T}$ of the model will be a probability of belonging to each sleep stage.

The presented deep learning approach is a hierarchical sequence-to-sequence classifier. RobustSleepNet is almost entirely based on SimpleSleepNet \cite{guillot2019dreem}, which was itself built on top of SeqSleepNet \cite{SeqSleepNet}. However, we tweaked the epoch encoder using the attention layer described in \cite{bahdanau2014neural}. This rather small design change has great importance since it solves the input-shape incompatibility described in part \ref{introduction}. In this way, it enables RobustSleepNet to handle an arbitrary number of input channels $C_\mathcal{D}$ by recombining them into a fixed number of channels. It allows training on any montage and multiple datasets at the same time. One other collateral property of using attention is handling input channels in any order (channel-order invariance).

\begin{figure}
\vspace{-1em}
    \begin{center}
    \includegraphics[width= 1 \linewidth]{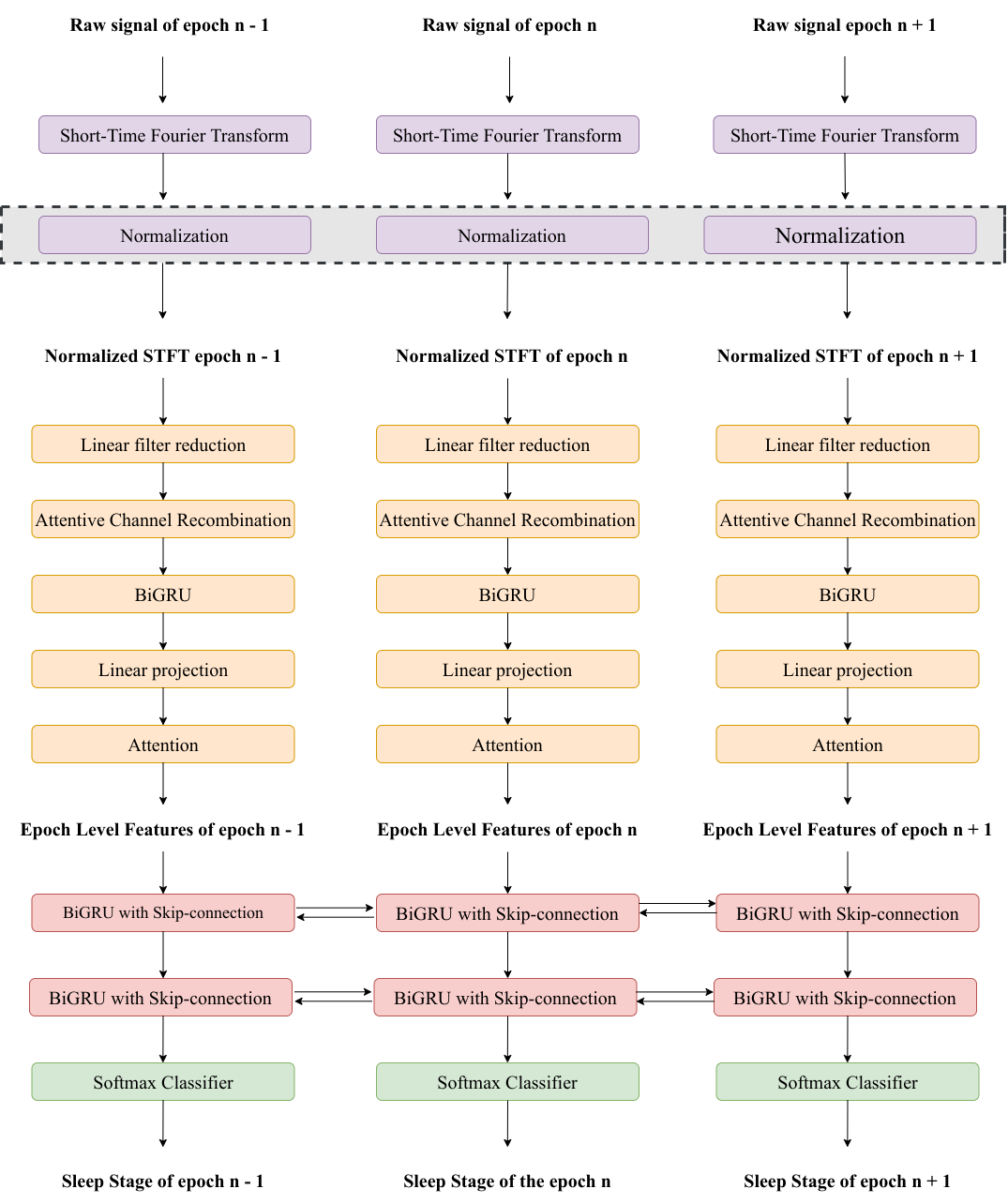} 
    \caption{The architecture of RobustSleepNet when the temporal context is 3. The first block is in purple (spectrum and normalization), the second block is in yellow (epoch encoder), the third in red (sequence encoder), and the fourth in green (classifier). The grey area illustrates that we normalized using all the epochs from the temporal context.}
    \label{fig:schema_du_modele}
    \end{center}
    \vspace{-2em}
\end{figure}

The architecture is summarised in Fig \ref{fig:schema_du_modele}. We implemented RobustSleepNet in PyTorch \cite{pytorch}. The code, trained models and results, are publicly available at \url{https://github.com/Dreem-Organization/RobustSleepNet}.

RobustSleepNet is composed of four main blocks. The first block processes the raw signal of one epoch $\mathbf{X} \in \mathbb{R}^{C \times L}$, and computes its normalized short-time Fourier transform (STFT) in $\mathbb{R}^{C \times F_{fft} \times L_{fft}}$. The second block encodes a segment from $\mathbb{R}^{C \times F_{fft} \times L_{fft}}$  into a vector $\mathbb{R}^P$. The third block uses the output vectors from $T$ epochs. It encodes the resulting $\mathbb{R}^{P \times T}$ into another sequence of smaller vectors in $\mathbb{R}^{Q \times T}$. The last block is a classifier that maps the sequence of vectors from $\mathbb{R}^{Q \times T}$ to a probability of belonging to each class in $[0, 1]^{5 \times T}$.

% bloc 1
\subsubsection{Signal normalization}
Given raw signals from $X \in \mathbb{R}^{C \times L}$, we computed the Fourier transform on $N_{fft}$ points every $N_{stride}$ points using a Hamming window. Hence, the STFT of the signal is $\mathbf{X}_{fft} \in \mathbb{R}^{C \times F_{fft} \times L_{fft}}$ where $F_{fft} = N_{fft} //2 +1 $ and $L_{fft} = (L - N_{fft}) // N_{stride}$.
Then, as in \cite{Optimal_transport_chambon}, we normalized each slice across time for each signal to have zero-mean and unit variance. Finally, the mean and variance were computed over the $T$ epochs independently for each channel and frequency bin.

% bloc 2
\subsubsection{Epoch encoder}
The segment encoder was inspired by SeqSleepNet \cite{SeqSleepNet} and SimpleSleepNet \cite{guillot2019dreem}. First, the $F_{fft}$ frequency bins were linearly projected onto smaller $F_{red}$ bins using a matrix $\mathbf{W_f} \in \mathbb{R}^{F_{red} \times F_{fft}}$  and a bias vector  $b_f \in \mathbb{R}^{F_{red}}$. The output was called $\mathbf{X}_{red} \in \mathbb{R}^{C \times F_{red} \times L_{fft}}$. 

Then, we recombined the channels into a fixed number using the attention introduced in \cite{bahdanau2014neural}. This recombination is a critical difference in RobustSleepNet compared to previous sleep staging models as it tackles the input-shape issue described in part \ref{introduction}. It enables the use of any number and any combination of input channels; moreover, the channel order can be arbitrary. An attention layer is called an attention head, and it recombines all the $C$ channels into a single channel — RobustSleepNet can apply several attention heads in parallel.

Each attention head had two weights matrices $\mathbf{W_{ctx}} \in \mathbb{R}^{F_{red}, K_1}, b_{ctx} \in \mathbb{R}^{K_1}$ and $\mathbf{W_{v}} \in \mathbb{R}^{K_1,1}$. We define $x \in \mathbb{R}^{C \times F_{red}}$ a window of the reduced spectrogram $\mathbf{X}_{red}$. The same operation was performed for all the windows.

\begin{align*}
u & = tanh(x \mathbf{W_{ctx}} +  b_{ctx}) & \in \mathbb{R}^{C \times K_1} \\
uv & = u.\mathbf{W_{v}} & \in \mathbb{R}^{C} \\
\alpha & = \frac{exp(uv)}{\sum{exp(uv)}} & \in \mathbb{R}^{C} \\
x_{head} & = \alpha^T.x & \in \mathbb{R}^{F_{red}}
\end{align*}

The recombined channel was defined as the average of the input channels weighted by their importance score $\alpha$ and this operation was performed independently for each of the $N_{heads}$ attention heads. We then concatenated their output, i.e., recombined the original $C$ channels into $N_{heads}$ channels. $K_1$ was the context size of the attention layers. The output of the layer was of the shape $\mathbf{X}_{att} \in \mathbb{R}^{N_{heads} \times F_{red} \times L_{fft}}$. 

We flattened the result to get $\mathbf{X}_{flat} \in \mathbb{R}^{(N_{heads}.F_{red}) \times L_{fft}}$. We fed $\mathbf{X}_{flat}$ to a bidirectional GRU \cite{GRU2014} with $H_1$ hidden units to get $\mathbf{X}_{gru} \in \mathbb{R}^{2.H_1 \times L_{fft}}$. During training, we applied dropout \cite{dropout} before and after the GRU with a probability $p_1$.

We projected each output of the GRU linearly into a vector $X_{proj} \in \mathbb{R}^{P \times L_{fft}}$. Similarly to \cite{SeqSleepNet}, we then reduced $X_{proj}$ along the time axis with an attention layer to $X_{features} \in \mathbb{R}^{P}$ . We denoted $K_2$ the context size of this attention layer.

% bloc 3
\subsubsection{Sequence encoder}
The first two blocks encoded a sequence $\mathbf{Z} \in \mathbb{R}^{ C \times L \times T}$ into a sequence $\mathbf{Z}_{features} \in \mathbb{R}^{ P \times T}$
The sequence encoder was a two-layer bidirectional GRU with skip-connection \cite{supratak2017deepsleepnet} \cite{guillot2019dreem} encoding $\mathbf{Z}_{features}$ into $\mathbf{Z}_{encoded} \in \mathbb{R}^{ Q \times T}$. $Q = 2 \times H_2$, where $H2$ denotes the number of hidden units in each layer and $p_2$ the dropout applied after each layer.

% bloc 4
\subsubsection{Classifier}
The classifier was a softmax layer that mapped $\mathbf{Z}_{encoded}$ to $\hat{\pi} \in [0, 1]^{5 \times T}$ containing the probability for each epoch of the sequence to belong to each sleep stage.

\subsection{Loss function}
As in \cite{SeqSleepNet}, the loss was computed over all the temporal context. Considering an input sequence $\mathbf{Z} \in \mathbb{R}^{ C \times L \times T}$, its ouput $\hat{\pi} \in [0, 1]^{5 \times T}$ and the corresponding ground truth labels $y \in \{ 0, 1 \}^{5 \times T}$ the loss was expressed as:
$$\mathcal{L}(\mathbf{Z},y) = -\frac{1}{T} \sum_{i= 0}^{T} y_{i}  \boldsymbol{\cdot} log(\hat{\pi}_{i})$$

\subsection{Dealing with varying channel size during training in DT}
We showed how the multi-head attention part of the network could deal with varying channel sizes $C_\mathcal{D}$. However, it is technically more convenient to deal with a fixed number of channels for each batch of observations during training.  To do so while keeping a mix of observations from several datasets in each batch, we will define in the following section a way to select a fixed $C_{batch}$ number of channels for a batch.

$C_{batch}$ was sampled for each batch in $C_{batch} \in \{1, .., C_{max}\}$ with $C_{max} = \max_{d \in \mathcal{D}}( C_{d} )  $ following: $$\mathbf{P}(C_{batch} = n) = \frac{1}{\sum\limits_{C=1}^{C_{max}} \frac{n}{C}}$$

This encouraged smaller $C_{batch}$ sizes, which prevented the model from relying on too many channels to make a prediction.
Then, we randomly sampled $C_{batch}$ channels for each observation with replacement if $C_{batch} > C_{\mathcal{D}}$. We used only a subset of the channels. This sampling method acted as regularization because it reduced the dependency of the network on specific channels and channel order.

\subsection{Inference}
To infer the sleep stages on a dataset $\mathcal{D}_{target}$, we evaluated the trained model on each subject with a stride of 1. For each epoch $\mathbf{X} \in \mathbb{R}^{C \times L}$ of a record, we obtained $T$ predictions $\hat{\pi}_{i}$ due to the temporal context and the stride. As in \cite{SeqSleepNet} and \cite{guillot2019dreem}, the $T$ estimates were aggregated using a geometric average. $$\tilde{\pi} = \text{exp}\left( \frac{1}{T} \sum_{i = 0}^{T} log(\hat{\pi}_{i})\right)$$, and the predicted sleep stage used for evaluation was $$\tilde{y} = \text{argmax}_{\; j \in [\![0,5]\!]} \tilde{\pi}_j $$

\section{Experiments and evaluation}

\subsection{Experimental Setup}

\subsubsection{Datasets}
\begin{table*}[t]
\centering

\scalebox{0.95}{%
\setlength\tabcolsep{4pt}
\begin{tabular}{|c|c| c| c |c c c |c c c c c c c c|c|c|c|c|} 
 \hline
 \setlength\tabcolsep{4pt}
  & & DOD-O & DOD-H & \multicolumn{3}{c|}{MASS} & \multicolumn{8}{c|}{Sleep-EDF} & MrOS & MESA & SHHS & CAP\\ 
  \setlength\tabcolsep{4pt}
 &  & &  & SS1 & SS3 & SS1+SS3 & \multicolumn{2}{c}{SC 39} & \multicolumn{2}{c}{SC 153}  & \multicolumn{2}{c}{ST}  & \multicolumn{2}{c|}{Full}  &  &  &  & \\ 
 
 & Model  & &  & & & & (1) & (2) & (1) & (2) & (1) & (2) & (1) & (2) &  &  &  & \\ 
 \hline
 \multirow{7}*{\rotatebox{90}{LFS}}
       & RobustSleepNet & \textbf{80.5} & 81.6 & \textbf{76.7} & 82.2 &\textbf{81.1}   &  79.1 & \textbf{78.8} & 76.3  & 75.4 & \textbf{78.6}  & \textbf{77.7} & 77.8 & \textbf{77.0}  & \textbf{74.1} &  \textbf{79.2} & 79.2 & \textbf{71.8} \\
           & XSleepNet2 \cite{phan2021xsleepnet} & - & -  & - & - & - & \textbf{80.9} & 78.7 & \textbf{78.7}  & \textbf{76.7} & - & -  & - & - & - & - & \textbf{82.3} & -\\
               & SimpleSleepNet \cite{guillot2019dreem} & 79.4 & \textbf{82.2} & - & 84.6 & - & 80.5 & - & -  & -   & - & - & \textbf{79.0} & -  & - & - & -  & - \\
            &  SeqSleepNet \cite{MultipleChannelTrasnferLearning}\cite{guillot2019dreem} \cite{phan2021xsleepnet}& - & - & - & 83.2 & - & 79.7 & 74.2 & 78.2 & 75.7  & - & 74.8 & 78.5 & - & -& - & 80.7& - \\

& DeepSleepNet \cite{guillot2019dreem} \cite{SEO2020102037}& 77.3 & 82.1  &- & \textbf{85.2} &- & 80.4 & - &-  & - & - & -  & 78.1 & - & - & - & 78.5 & -\\
        & GraphSleepNet \cite{GraphSleepNet} & - & - & - & 84.1 & - & - & - & -  & -   & - & - & - & -  & - & - & -  & - \\
& IITNet \cite{SEO2020102037} &  - & - & - & 80.5 & - & 77.6  & - & - & -  & -  & -& - & - & - & - & 79.8 & -\\

    & UTime \cite{perslev2019utime}& - & -& - & -& - & 79.0 & -  & 76.0 &  -& - & - & -  & - & - & - & - & 68 \\
 & CNN \cite{varela} & - & -& - & - & - & - & -& - & - &  - & - & -  & -  & -  & - & 76.0 & - \\
 & CNN \cite{Sors2018ACN} & - & -& - & - & - & - & - &  & - & - & -  & -   & - & - & - &  78.0 &  \\
 \hhline{|=|==================|} 
 \multirow{5}*{\rotatebox{90}{FT}} 
 & RobustSleepNet & \textbf{82.7} & \textbf{85.1} & \textbf{79.7} & \textbf{84.0}  & \textbf{82.5}& \textbf{81.7} & \textbf{81.3}  & \textbf{77.9}& \textbf{77.0}  & \textbf{81.0} & \textbf{81.0} & \textbf{79.6} & \textbf{78.3}  & \textbf{75.6} &  \textbf{79.5} & \textbf{80.0} & \textbf{73.8}\\
     & XSleepNet2 \cite{phan2021xsleepnet} & - & -  & - & - & - & - & 80.8 & -  & - & - & -  & - & - & - & - & - & -\\
 & SeqSleepNet (EEG)\cite{SingleChannelMismatch} \cite{MultipleChannelTrasnferLearning} & - & -& - & - & - & - & 80.0  & - & -& - & 77.5  & - & - & - & -& -  & - \\
    &  SeqSleepNet (EEG-EOG) \cite{MultipleChannelTrasnferLearning} & - & - & - & - & -& - & 77.7 & -& -& -  & 76.7  & -& -& - & - & - & - \\
    &  DeepSleepNet (EEG) \cite{MultipleChannelTrasnferLearning} & - & - & - & - &  -& - &78.8 & -& - & - & 77.5  & - & - & - & - & - & - \\
        &  KL-reg \cite{KLPersonalization} & - & -& -& - & - & - & - & - & 73.0   & - & - & - & - & - & - & - & - \\
   \hhline{|=|==================|} \multirow{6}*{\rotatebox{90}{DT}} & RobustSleepNet & \textbf{79.2} & \textbf{84.4} & \textbf{75.6} & \textbf{80.8} &  \textbf{78.7}  & \textbf{79.1} & \textbf{78.2} & \textbf{73.8} & \textbf{73.1} & \textbf{79.1} & \textbf{79.1} & \textbf{75.2} & \textbf{74.4} & \textbf{72.6} &  \textbf{74.9} & \textbf{76.3} & \textbf{64.9}\\
 &  SeqSleepNet (EEG)\cite{SingleChannelMismatch} & - & - & -&  - & -& -& 74.6 & - & - & -& 75.6  & - & - & - & - & -& - \\
 &  SeqSleepNet (EEG-EOG)\cite{MultipleChannelTrasnferLearning}& - & -& -& - & - & - & 62.1 & - & - & - & 64.2 & - & - & -& -& -& - \\
  &  DeepSleepNet (EEG) \cite{MultipleChannelTrasnferLearning} & - & -& -& - & - & - & 66.9 & - & - & - & 61.3 & - & - & -& -& -& - \\
 &  SimpleSleepNet \cite{guillot2019dreem} & 43.4 & 72.0 & - & -& - & - & - & -& - & - & - & - & - & - & -& -& -\\
    & Optimal Transport \cite{Optimal_transport_chambon} & - & - & - & 60.0& -& - & - & - & - & - & - & - & - & -& -& - & - \\

 \hline
\end{tabular}
}
\centering
\caption{RobustSleepNet benchmark on the three considered learning settings (LFS, FT, DT) for the eight datasets. Average macro-F1 across epochs is reported. Performance of the literature is reported when available. Bold font highlights the highest values for each dataset and each setting. (1) refers to the setup used in \cite{supratak2017deepsleepnet} \cite{perslev2019utime} and (2) corresponds to the in-bed setup described in \cite{MultipleChannelTrasnferLearning}.  The epoch-wise macro F1 on DOD-O and DOD-H \cite{guillot2019dreem} was computed using the publicly available \href{https://github.com/Dreem-Organization/dreem-learning-evaluation}{results}.  For \cite{MultipleChannelTrasnferLearning}, we report the performance when using EEG for SeqSleepNet and EEG-EOG for SeqSleepNet and DeepSleepNet.}

\label{tab:main_table}
\vspace{-2em}
\end{table*}

We consider eight sleep staging datasets with variable demographics and PSG montages. Demographics of the datasets are reported in Table \ref{tab:dataset_desctiption}.

\textbf{Dreem Open Datasets Obstructive (DOD-O):} \cite{guillot2019dreem} Public dataset scored by five sleep-experts according to the AASM standards. DOD-O is composed of 55 subjects diagnosed with sleep apnea. EEG, EOG, EMG, and ECG derivations are available. 
Unless otherwise stated, we use the consensus of the five scorers as the ground-truth hypnograms for DOD-O. \cite{guillot2019dreem} describes the procedure to build the consensus.

\textbf{Dreem Open Datasets Healthy (DOD-H):} \cite{guillot2019dreem} Public dataset scored by five sleep-experts according to the AASM standards. DOD-H is composed of 25 nights from healthy subjects without any sleep-related disorder or medications. EEG, EOG, EMG, and ECG derivations are available. 
Unless otherwise stated, we use the consensus of the five scorers as the ground-truth hypnograms for DOD-H\cite{guillot2019dreem}.

\textbf{Montreal archive of sleep studies (MASS) :}\cite{MASS} We consider only the SS1 and SS3 cohorts from MASS. We exclude SS2, SS4, and SS5, which use the R\&K scale with 20 seconds epochs. Both cohorts were scored according to the AASM standards. The SS1 cohort comprises 53 nights from elderly subjects (some of whom have sleep-related disorders) and the SS3 cohort of 62 nights from healthy subjects. A complete PSG montage was used. We report performances on the two subsets. 

\textbf{Sleep-EDF dataset :} \cite{SleepEDF} The Sleep-EDF dataset is composed of two studies, Sleep-Telemetry (ST) and Sleep-Cassette (SC). ST includes 44 nights from 22 subjects with trouble falling asleep. Half of the nights were done after taking Temazepam, and the others after taking a placebo. SC includes 153 nights from 82 subjects who do not take sleep-related medications. 
Both studies are scored used the R\&K standards\cite{rechtschaffen1973manual}, similarly to other works \cite{guillot2019dreem} \cite{SeqSleepNet} \cite{perslev2019utime}, we merge stages S3 and S4 into N3. 
For both studies, we use the  FPZ-Cz,  Pz-Oz, and the  EOG derivations. To compare with the literature, we consider the following subsets at evaluation time: SC-39 (the first 39 records from the SC study), SC-153 (all the 153 records from the SC study), and ST (the 44 records of the ST study).
In previous works, two evaluation setups were described. In the base setup, only epochs in-between 30 minutes before the first non-wake epoch and 30 minutes after the last non-wake epoch are used \cite{supratak2017deepsleepnet} \cite{perslev2019utime}. The in-bed setup where only epochs between lights-off and lights-on time are considered \cite{sleep_edf_processing} \cite{MultipleChannelTrasnferLearning}. 

\textbf{MrOS :}  \cite{MrOS1} \cite{MrOS2} \cite{MrOS3}  MrOS is a child study analyzing the link between osteoporotic Fractures and sleep. The nights were scored by sleep experts from six different sleep centers according to the AASM standard.  We consider the 586 nights from the second Sleep Study MrOS-2 with EEG, EMG, and EOG derivations. 

\textbf{The Multi-Ethnic Study of Atherosclerosis (MESA) :} \cite{NSSR1} \cite{NSRR2} The Multi-Ethnic Study of Atherosclerosis dataset contains 2,237 nights from subjects likely to suffer from cardiovascular diseases aged from 54 to 95. The nights are scored according to the AASM standard by different scorers from six sleep centers. The subjects are not screened for sleep-related pathologies and may suffer from various pathologies. The PSG montage contains Fz-Cz, Cz-Oz, C4-M1, EMG, EOG, and ECG derivations.

\textbf{Sleep Health Hearth Study (SHHS) :}  \cite{SHHS1} \cite{SHHS2} The Sleep Health Hearth Study is composed of two multi-center polysomnographic studies of subjects aged 40 years and older. The nights are scored according to the R\&K standard \cite{rechtschaffen1973manual} by different scorers from multiple sleep centers. We merge stages S3 and S4 into N3. We consider the second study SHHS-2 composed of 2651 nights with two EEG, one EOG, and one EMG derivation. In the literature most publications use 70\% of the dataset for training and the remaining 30\% for evaluation without further details.

\textbf{The Cyclic Alternating Pattern dataset (CAP):} The CAP dataset is composed of 108 nights, with 92 pathological records from subjects with a REM behavior disorder, periodic leg movements, nocturnal frontal lobe epilepsy, sleep apnea, narcolepsy, and bruxism. The dataset is scored according to the R\&K standard \cite{rechtschaffen1973manual}. We merge Stages S3 and S4 into N3. The nights are recorded and scored in a single sleep center and each PSG contains at least 3 EEG channels, EOG, EMG, and ECG derivations. The records nfle25 and nfle33 are excluded because of missing channels.

\subsubsection{Preprocessing}
We preprocess the data record-wise for every dataset. First, we apply an IIR band-pass filter between 0.2 and 30Hz to ensure the frequency content consistency between the datasets. This frequency range contains the information needed for sleep scoring. We then resample the signals to 60Hz using polyphase filtering from the scipy library. We scale each record to have a unit-interquartile range and zero-median to ensure similar scale across datasets as in \cite{perslev2019utime}. To ensure the stability of the training, we clip values that are greater than 20 times the interquartile range. 

\subsubsection{Model parameters}
We consider the five AASM sleep stages computed over segments of $L=1800$ points corresponding to 30 seconds sleep epochs at 60Hz. Similarly to recent works\cite{SeqSleepNet}\cite{guillot2019dreem}, we set the temporal context $T$ to 21 sleep-epochs (630 seconds). We set $N_{fft}$ to 128 and $N_{stride}$ to 60. We set $F_{red}$ to 32. The number of heads to $N_{heads}=4$, and the channel recombination attention to $K_1 = 30$. The GRU layer has $H_1 = 64$ hidden units and dropout set to $p_1=0.5$. The projection of the output of the GRU is done with $P=50$ parameters.
The second attention has a context size $K_2 = 25$. The sequence encoder has $H_2 = 50$ hidden units. We set the dropout $p_2=0.5$. With this configuration, RobustSleepNet has 180,343 parameters, which is similar to \cite{guillot2019dreem}, and an order of magnitude lower than other recent models \cite{supratak2017deepsleepnet}
 \cite{SeqSleepNet} \cite{perslev2019utime} \cite{phan2021xsleepnet}.

We train RobustSleepNet using backpropagation with the Adam optimizer and a learning rate of 1e-3 for DT and LFS and of 1e-4 for FT, momentum parameters $\beta_1=0.9$ and $\beta_2=0.999$ and a batch size of 32. We train the model for a maximum of 100 epochs with early stopping when the validation accuracy stops improving for more than five epochs. In practice, the stopping criterion was often reached within 20 epochs. We use the model with the best validation accuracy to evaluate the model. Training RobustSleepNet on 1000 nights takes between 6 and 8 hours on an Nvidia GeForce RTX 2080Ti.  

\subsection{Benchmark} \label{sec:benchmark}
In this first experiment, we benchmark RobustSleepNet using the three training settings described in section \ref{sec:problem_def} on each dataset. We are interested in comparing the DT performance and FT performance of RobustSleepNet with its LFS performance. The latter being the one commonly used in the literature. We also provide values from the literature as references.

In the DT setting, we train and validate the model on seven datasets out of the eight datasets. The seven datasets are the source datasets. We evaluate the model on the remaining target dataset. In the DT setting, we use 70\% of the training records from the source datasets to train the model, and we use the 30\% remaining for validation.
We select a subset of 250 records for the larger datasets (MESA, MrOS, SHHS) to have a similar proportion of records from each of the datasets during training. We select the 250 records with the highest macro F1 in the LFS setting. This selection discards records with poor quality or scoring.

In the LFS and FT settings, we use k-folds to evaluate the model; hence, we can predict a hypnogram for each record of the dataset correctly, i.e., without using it during training. We indicate the number of folds in \ref{tab:split}; as there are several nights from the same subject in Sleep-EDF, we group all the nights of one subject in the same fold to avoid subject bias. In the FT setting, we use the final weights of RobustSleepNet obtained in the DT setting to initialize the model. 
We use fewer folds than in previous works due to computational complexity; as a result, the model is trained using fewer training examples. This setup could disadvantage LFS and FT setting models, especially for the smallest datasets, because fewer data are available during training than in the literature.
As described in \ref{part_2_evaluation}, we report epoch-wise F1 for consistency with the literature. For RobustSleepNet, the F1 provided is an average over three independents training runs. This average mitigates training randomness. For other approaches, we report results from the original papers. To our knowledge, they are computed on a single run and do not provide uncertainty bounds.
\begin{table*}[t]
\centering

\scalebox{1}{%
\setlength\tabcolsep{4pt}
\begin{tabular}{|c | c| c |c c c |c c c c |c|c|c|c|} 
 \hline
 \setlength\tabcolsep{4pt}
  & DOD-O & DOD-H & \multicolumn{3}{c|}{MASS} & \multicolumn{4}{c|}{Sleep EDF} & MrOS & MESA & SHHS & CAP\\ 
  \setlength\tabcolsep{4pt}
 & &  & SS1 & SS3 & Full & SC39 & SC153  & ST  & Full &  &  &  & \\ 
 LFS Folds & 5  & 5  & 5 & 5 & 5 & 5 & 5 & 5 & 5 & 3 & 3 & 3 & 5 \\
 FT Folds & 5 & 5   & 5 & 5 & 5 & 5 & 5 & 5 & 5 & 3 & 3 & 3 & 5\\
 Folds in litterature & 10 \cite{guillot2019dreem} & 25 \cite{guillot2019dreem}  & NA \cite{supratak2017deepsleepnet} & 31 & NA & 20 \cite{SeqSleepNet} & 10 \cite{phan2021xsleepnet} & 11 \cite{MultipleChannelTrasnferLearning} & 10 \cite{guillot2019dreem} & NA & NA & 70 \% Train / 30 \% Test  \cite{phan2021xsleepnet} & 5 \cite{perslev2019utime}\\
 \hline
\end{tabular}
}
\centering
\caption{Number of folds used to train RobustSleepNet in the FT and DT setup. For reference, we report the number of folds used in the literature for each dataset in the last row.}

\label{tab:split}
\end{table*}

We report the results of RobustSleepNet and from the literature for the three settings in Tab. \ref{tab:main_table}. 
In the LFS setting, RobustSleepNet performs well on most datasets. It performs on par with recent models from the literature  \cite{SeqSleepNet}\cite{guillot2019dreem}\cite{phan2021xsleepnet}\cite{supratak2017deepsleepnet} . The FT setting provides an average 2.5\% relative increase of macro F1 when compared to the LFS setting. In this setting, RobustSleepNet outperforms other methods from the literature on all the datasets except on MASS SS3 and SHHS. On SS3, DeepSleepNet has a slightly higher F1 while XSleepNet has a higher F1 on SHHS (but is only evaluated on 30\% of the subjects). The FT setting also provides a 5.1\% relative F1 increase compared to DT. The improvement is more significant on CAP, MrOS, MESA, and SHHS and more minor on DOD-H, SC39, and ST.

In the DT setting, RobustSleepNet performs much better than previous attempts from the literature. Interestingly, it outperforms the LFS F1 on DOD-H, SC39, and ST. It also reaches the same level of performance as in the LFS setting on DOD-O and MASS SS1. On MrOS, MESA, and SHHS, it has a 3\% to 4\% lower accuracy. On CAP, the F1 drop by 7\% between LFS and DT.

\subsection{Influence of the experimental parameters in DT}

\subsubsection{Number of training records} 
\begin{figure}[t]
\vspace{-1em}
    \begin{center}
    \includegraphics[width=\linewidth]{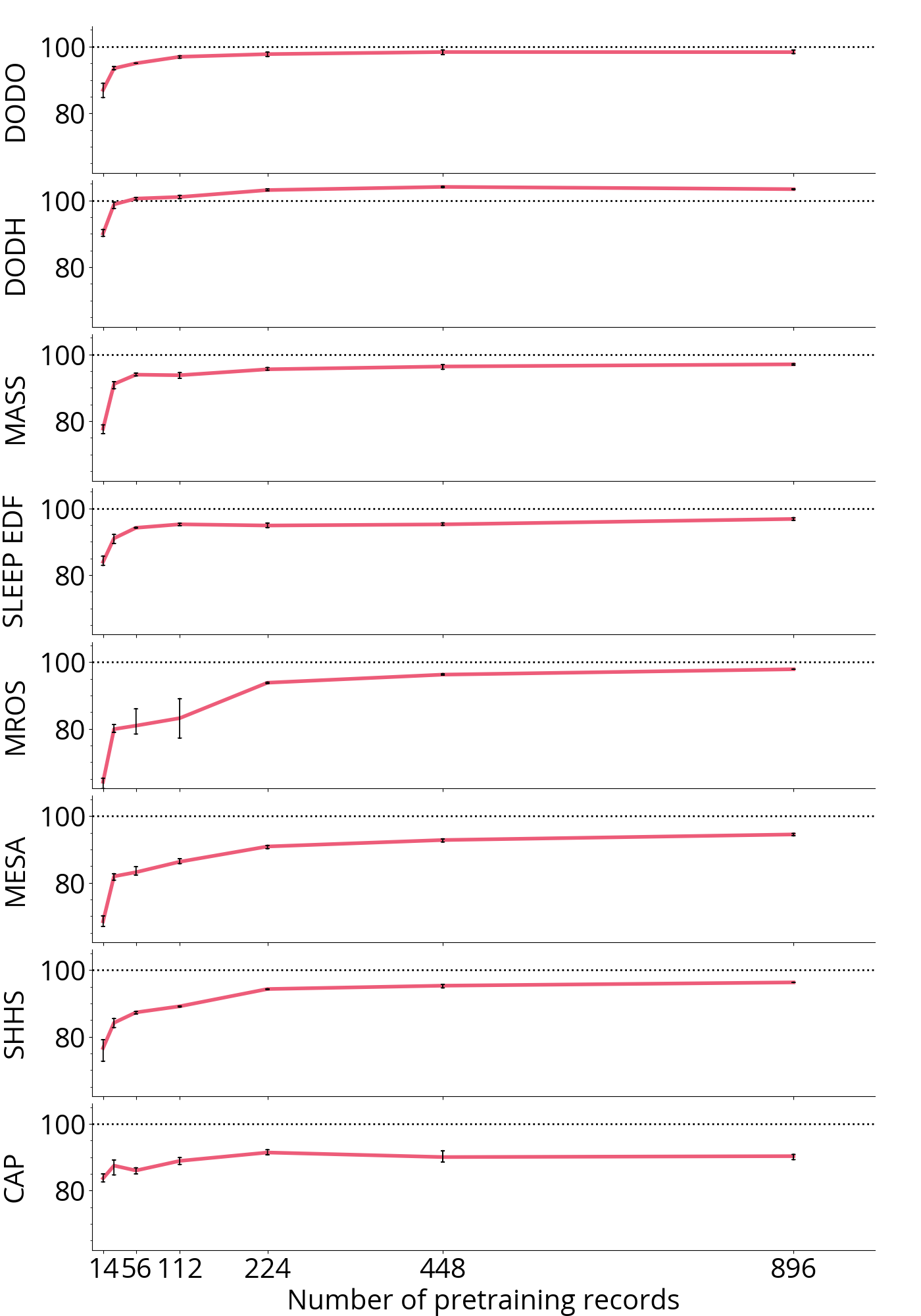} 

    \caption{Evolution of the F1 of RobustSleepNet in DT depending on the size of the training set. We report the performance as the percentage of the LFS performance achieved by RobustSleepNet. The error bars report the min and maximum F1 over three repetitions. The horizontal line at 100 represents the performance achieved in the LFS setting.}
    \label{figure:number_of_records}
    \end{center}
\vspace{-2em}
\end{figure}

We assess how the performance of RobustSleepNet in the DT setting evolves depending on the number of records used for training. We incrementally increase the number of records used for training from two records per dataset (14 records) to 128 records per dataset (896 records). We sample the same number of records per dataset when possible. When a dataset contains fewer records than required, we use all the records from the dataset. The remaining required records are sampled from the other datasets. Figure \ref{figure:number_of_records} shows the performances on the complete target dataset. We report the performance as the percentage of the LFS performance achieved by RobustSleepNet.

The performance sensibly increases within the first few records. The results variability also decreases with the number of training records, and the F1 increases slightly with every additional record. Interestingly, we reach the LFS baseline within the first few records on DOD-H.

\subsubsection{Number of attention heads}: To assess the importance of having multiple attention heads on performance, we train and evaluate a RobustSleepNet model with up to 16 attention heads. We report the average F1 for over the eight datasets in \ref{tab:attention_heads}

\begin{table}

\centering
\begin{tabular}{|c|c|c|c|c|c|} 
\hline
Attention Heads & 1 & 2 & 4 & 8 & 16 \\
Avg. F1 &  75.3 & \textbf{75.8} & \textbf{75.8} & 75.4 & 75.1  \\

\hline
\end{tabular}
\caption{Avg. F1 over the 8 datasets in the DT settings depending on the number of attention heads in RobustSleepNet}.
\label{tab:attention_heads}
\end{table}

\subsubsection{Influence of the channels in the target dataset}
\begin{figure}
\vspace{-1em}
    \begin{center}
    \includegraphics[width=\linewidth]{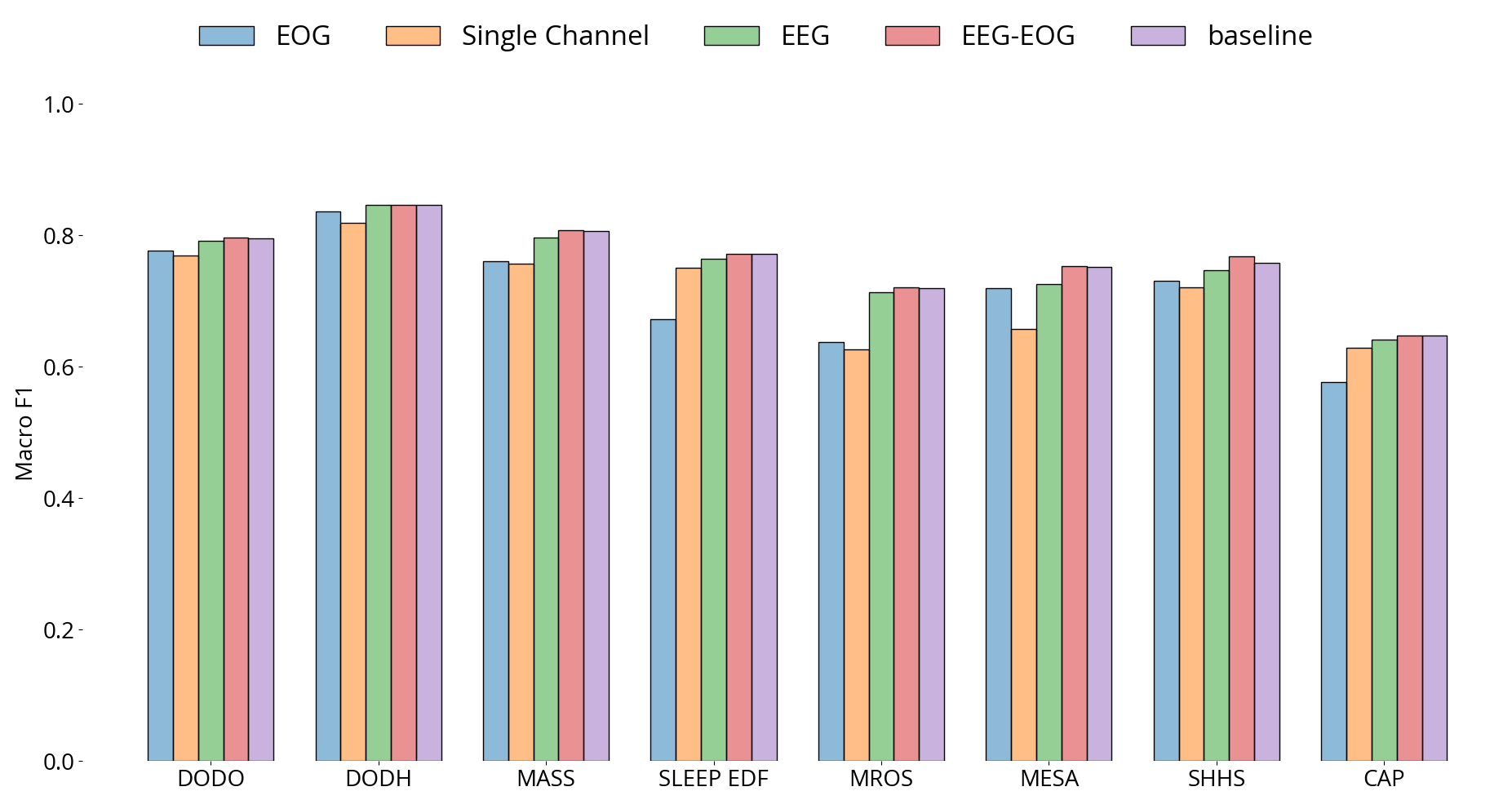} 

    \caption{DT Performance of RobustSleepNet using various PSG montages. All the available EEG and EOG for EEG + EOG, a single EEG Channel for single EEG. The baseline contains the channels from all the montages for this dataset.}
    \label{figure:scorers_transfer}
    \end{center}
    \vspace{-1em}
\end{figure}

: To assess the impact of removing specific channels on performance, we evaluate RobustSleepNet in the DT setting a using montage with different channels: \textit{Single EEG} (Best performing EEG channel), \textit{EEG} (all the available EEG channels), \textit{EOG} (all the available EOG channels), \textit{EEG+EOG} (all the EEG and EOG channels). The baseline montage uses all the available derivations described in Tab. \ref{tab:dataset_desctiption}. Its results are reported in Tab. \ref{tab:main_table}. Results are presented in Figure \ref{figure:scorers_transfer}.

The EEG combined with the EOG performs along with the baseline on every dataset. The average F1 drop is 1\%. When using a single EEG channel, the drop is around 5\% on average. It is around 10\% on MESA and MrOS and around 3\% on the other datasets.

\subsection{Influence of Interscorer variability on transfer learning}
A study by the AASM on 2500 nights showed an average inter-rater agreement of 82.6\% \cite{Rosenberg2014TheEvents} for sleep stage classification. We assess the impact of the inter-rater variability when training and evaluating RobustSleepNet using DOD-O and DOD-H. These two datasets include scoring from five scorers from different sleep centers. We rank the scorers based on their agreement with the other scorers \cite{10.1093/sleep/zsaa097} \cite{guillot2019dreem}. The scorer with the highest (resp. worse) agreement is ranked first (resp fifth).

\subsubsection{Influence on training}
We train RobustSleepNet based on the annotations of each scorer from DOD-O or DOD-H. We also train one model on the consensus of the four best scorers \cite{guillot2019dreem}. We then evaluate each model using the DT setting on the other datasets. We report the average F1 over all the datasets in Table \ref{tab:scorer_performances_source} for each model.

\begin{table}
\vspace{1em}
\centering
\begin{tabular}{|c|c|c|c|c|c||c|} 
 \hline
& \multicolumn{5}{c||}{Scorers' rank } & Consensus \\
Source Dataset & 1 & 2 & 3 & 4 & 5 &  \\ \hline
DOD-O &  \textit{65.3} & 65.1 & 64.1 &57.7 &58.7 & $\textbf{66.2}$\\
DOD-H &  59.8 & 58.6  & 59.5 & \textit{61.7} & 53.8 & $\textbf{61.3}$\\

  \hline
\end{tabular}
\caption{RobustSleepNet is trained on DOD-O or DOD-H, using annotations of a single scorer or the consensus of scorers. Scorer ranking and consensus follow \cite{guillot2019dreem}. The reported F1 is the average F1 obtained when evaluating the other seven datasets of this study. }
\label{tab:scorer_performances_source}
\vspace{-1em}
\end{table}

The choice of scorer implies a large variability in the final performance. Training the model on the consensus of scorers instead of an individual scorer increases the average F1 by 4\% when DOD-O is the source dataset and 2.5\% when DOD-H is the source dataset. The reported F1 values are correlated to the scorer reliability. For instance, the models trained on the annotations from the worst scorer on DOD-O and DOD-H also have the lowest F1.

\subsubsection{Influence on evaluation}
We perform a reverse experiment to assess the influence of inter-scorer variability on the evaluation of a model. We use the model trained in DT for DOD-O and DOD-H. We evaluate by comparing each of the individual scorers and to the consensus. The F1 is reported in Table \ref{tab:scorer_performances_target}. The values for the consensus are logically the ones reported in \ref{tab:main_table}.

\begin{table}

\centering
\begin{tabular}{|c|c|c|c|c|c||c|} 
\hline
& \multicolumn{5}{c||}{Scorers' rank} & Consensus \\
Target Dataset & 1 & 2 & 3 & 4 & 5 &  \\ \hline
DOD-O &  77.4 & \textit{79.0} & 77.9 &70.7 & 68.9 & $\textbf{79.7}$ \\
DOD-H & \textit{81.2}  &79.3  & 79.1 & 75.9 & 71.6 & $\textbf{84.6}$ \\ 

\hline
\end{tabular}
\caption{RobustSleepNet trained in DT for DOD-O and DOD-H is evaluated on every individual scorer and the consensus of the scorers. Scorers' ranking and consensus follow \cite{guillot2019dreem}. The consensus values are also the ones reported in table \ref{tab:main_table}}
\label{tab:scorer_performances_target}
\vspace{-1em}
\end{table}

The performances of RobustSleepNet correlate with the scorer reliability. There are significant drops in the reported F1 of RobustSleepNet when we use the worse scorers as the ground truth. The consensus yields the highest test F1 on both DOD-O and DOD-H.

\subsection{Datasets generalization}
\begin{figure*}[t]
\vspace{-1em}
    \begin{center}
    \includegraphics[width=0.75\linewidth]{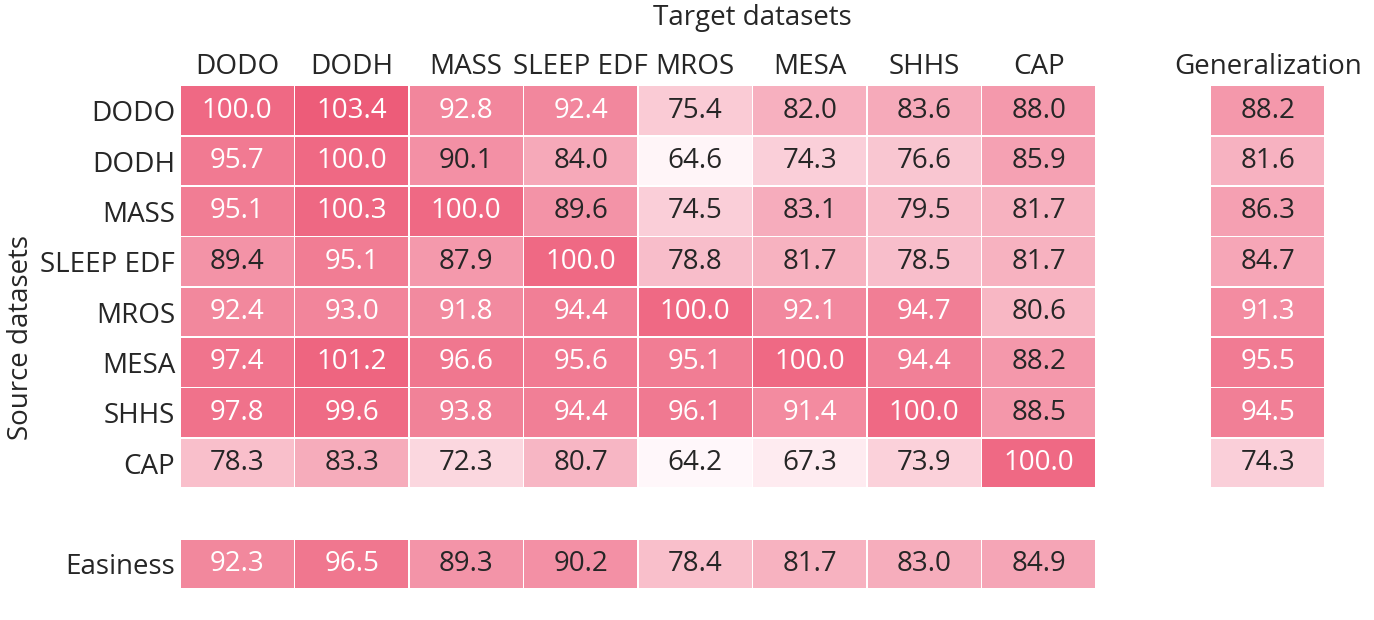}     \caption{Normalized F1, Easiness, and Generalization (defined in part \ref{section:easiness_and_gen}) of the 8 datasets. The easiness of a dataset $\mathcal{D}$ denotes how well a model trained on other datasets performs on $\mathcal{D}$ in DT. The generalization denotes how well a model trained on a $\mathcal{D}$ will generalize on other datasets in DT.}
    \label{figure:dataset_transferability}
    \end{center}
    \vspace{-0.5em}
\end{figure*}

We assess the generalization and easiness defined in part \ref{section:easiness_and_gen}) for each dataset. RobustSleepNet is trained on each of the individual datasets and evaluated on all the other datasets in DT. To limit the influence of the size of the dataset, we use the 250 records for MESA, MrOS, and SHHS selected in \ref{tab:main_table}.  We report the $F1_{norm}$ in-between all datasets. Using $F1_{norm}$ instead of F1, in that case, enables relative comparison between datasets. We also report easiness and generalization. The results are presented in Table \ref{figure:dataset_transferability}.

We observe three groups among the datasets. The first group is composed of the larger datasets MESA, MrOS, and SHHS. This group has the highest generalization, the lowest easiness, and the oldest subjects. They also present a large variety of pathologies. The second group of datasets is DOD-O, DOD-H, MASS, and Sleep-EDF. They have high easiness and low generalization values. These datasets include relatively young subjects and few pathologies (except sleep apnea) compared to the first group. The dataset CAP is an outlier with both low generalization and easiness.

\subsection{Number of finetuning records} 
\begin{figure}[t]
\vspace{-2em}
    \begin{center}
    \includegraphics[width=\linewidth]{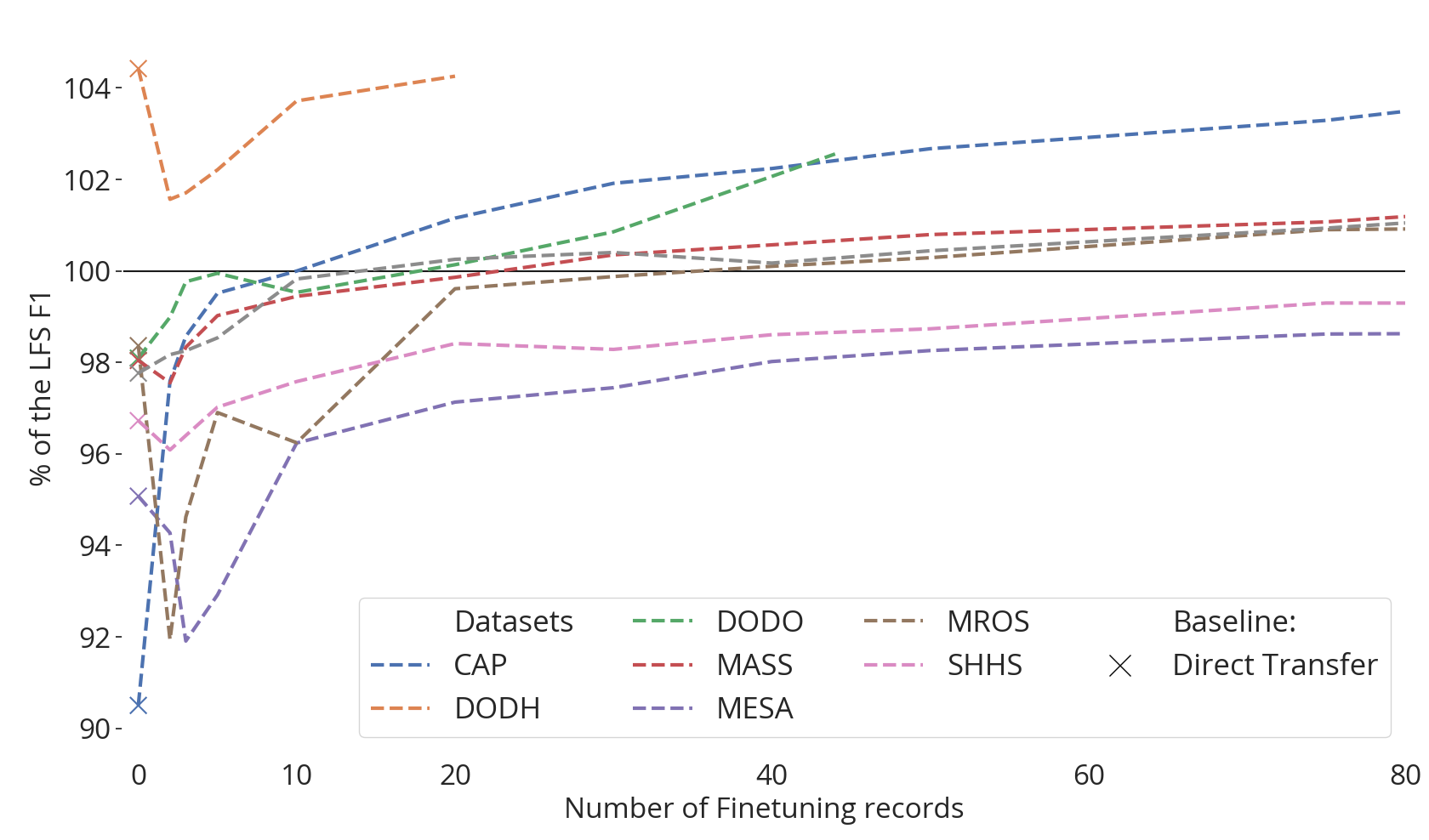} 
    \caption{FT performance relative to the number of training records. The F1 is expressed as the percentage of the LFS F1 from Table \ref{tab:main_table}. Hence, the LFS F1 is represented by the horizontal dashed line at 100. The performance when no finetuning records are used corresponds to DT performances.}
    \label{figure:learning_curve}
    \end{center}
    \vspace{-2em}
\end{figure}

We explore the evolution of the performance depending on the number of records used for FT. Starting from the DT model, we gradually increase the number $k$ of training records used for FT. As in part \ref{sec:benchmark}, 3 or 5 folds are used for the evaluation.
We randomly select the $k$ training records in each fold and use them for training and validation. We provide the results in Figure \ref{figure:learning_curve} as a percentage of the LFS performance from part \ref{sec:benchmark}.

We observe that the F1 decreases within the first few records before increasing for DOD-H, MrOS, and MESA. FT sensibly increases F1 when we use more than ten records. LFS F1 is reached within 10 to 40 records except for the two most extensive datasets, MESA and SHHS.

\section{Discussion}
RobustSleepNet can handle a variety of input signals using multi-head attentive recombination. This ability allowed for training the model on a corpus of eight datasets presenting a large variety of montages. Doing so, the model learned a large variety of demographics and input signals. In practice, it reached on average 97\% of the best LFS performance on each dataset, without using any information from the dataset. To our knowledge, this outperforms previous approaches from the literature \cite{MultiChannelTransfer} \cite{MultipleChannelTrasnferLearning} \cite{Optimal_transport_chambon} \cite{KLPersonalization}. Considering this DT performance, one can expect a pre-trained version of RobustSleepNet to perform well on unseen data. The only exception to this is the performance achieved on CAP in DT, which still presents a significant accuracy drop. However, we will discuss possible explanations as to why RobustSleepNet might underperform on that dataset. We have provided the code to use a pre-trained version of RobustSleepNet out of the box on an arbitrary PSG file. We trained this model on the eight datasets used in this study.

Increasing the number of records used for training in DT leads to a performance close to LFS performance. The performance is relatively high even with a few records, with on average 84\% of the supervised learning performance using only 14 records (2 records per source dataset) and 88\% using 28 records (4 records per source dataset). The fact that we use seven datasets with highly variable demographics to train the model might explain why we quickly reach that level of F1. One other clue to support this is that RobustSleepNet exceeds the LFS performance on DOD-H with only 56 records from other datasets. DOD-H contains only healthy young individuals; hence, it is likely to present low variability as suggested in \cite{guillot2019dreem}. The use of multiple records with high variability might be why it performs so well on this "easy" dataset.
While we demonstrated how the attention layers tackle the input-shape issue, the parametrization of that layer does not impact the performance much. All the attention heads use the same input, so they might tend to be similar to each other. Using several attention heads between 2 and 4 seems optimal on the considered datasets.
The channel ablation experiment highlights the ability of our model to perform well on various setups. In particular, the use of only a single EEG channel leads to exploitable results. The EOG also reached high performance, but this might be because of its almost frontal location means that it contains EEG information. In some cases, EOG was referenced to an EEG channel, which might bias that analysis. Single EEG performance is interesting for designing minimalist at-home PSG devices. A combination of EEG and EOG leads to the best results.

We quantified the impact of the inter-rater variability on training and evaluation. We show that training on a consensus of scorers is essential to achieve the best performance in DT. Several scorers reduce the noise due to inter-rater variability making RobustSleepNet more robust. We also observe a high variability on evaluation depending on the scorer used as the reference. RobustSleepNet generalizes well because we achieved the highest performances against the consensus. The high variations in the performance of individual scorers highlight the need to use such a consensus for a proper evaluation. When evaluating or training a model on a single scorer, one might be over-fitting on the scorer's way of classifying.

We assessed the individual contribution of each dataset of the corpus in DT by computing the normalized F1, the easiness, and the generalization metrics. The first observation one can glean is that easiness and generalization seem to be correlated and inversely correlated respectively with age. We interpret this observation in terms of variability: younger subjects present a lower level of variability than older subjects. Data from older subjects include information about young and old populations, whereas data from young subjects does not generalize to an older population. However, age is not the sole explanatory variable. If we consider DOD-O and CAP, we observe very different easiness and generalization even though subjects from both datasets are in the same range of age. DOD-O presents a lot more channels than CAP; hence it is likely to be more informative. Moreover, DOD-O was labeled by a consensus of several sleep experts, whereas a single scorer scored CAP. The consensus is probably boosting the generalization capacity of DOD-O, as discussed with the inter-rater variability experiment. Finally, DOD-O's subjects suffer from sleep apnea, whereas CAP's subjects suffer from various conditions that are less prevalent in the population. These factors could explain the low generalization value and easiness of CAP versus the relatively high for DOD-O. They also raise one limitation in the computation of generalization and easiness: the values are relative to the corpus of datasets selected. Eventually, the variability in the training datasets in terms of age, montage, PSG device, scoring, or pathology could all be explanatory variables to explain performance in DT. These conditions are met for this study's three most extensive datasets and have the highest generalization values. An interesting experiment would be, for each dataset, to train on an increasing number of datasets and assess the performance each time. However, this experiment is hard to perform and hard to plot due to the combinatory complexity it requires.

The performance of RobustSleepNet in LFS is on par with the literature models on each dataset. The proposed approach works well on small and medium databases but might be under-performing on the three largest. We link this to the relatively small number of parameters used on RobustSleepNet. However, using the fact that RobustSleepNet can be pre-trained on any dataset, we used FT and reached high performance on every dataset. In practice, there is no actual overhead cost to finetune a model instead of learning from scratch because one needs to retrain in every case. However, finetuning using only a few records leads to a drop in performance compared to the DT model for some datasets. That is why FT should be performed with sufficient records to improve performance and reach the LFS performance. Using around 10 to 40 records was enough for the small datasets of this study. Further increasing the number of records keeps improving performance slightly. Again, for the larger datasets of this study, more records were needed. However, in absolute, FT might be diminishing the generalization capacity of the model by overfitting on the specificity of the target dataset. One clue for that is that the largest datasets scored by several sleep experts in various sleep centers benefit from FT only with many records. This particularity suggests that they are hard to overfit on. CAP also benefits a lot from FT, supporting the previous observations that CAP is quite distinct from the other datasets in terms of variability. 
While not being the primary goal of this work, RobustSleepNet showed promising results in LFS and especially in FT. Future work should investigate its performance compared to state-of-the-art models more thoroughly in a comparative study such as the one realized in \cite{phan2021xsleepnet}.

One interesting axis of improvement for RobustSleepNet would be integrating the information related to the sensor positions, particularly the location of each EEG channel on the head. Indeed, the attention layer currently removes the information that could be beneficial in some specific cases. Integrating the position could allow RobustSleepNet to integrate better a variety of sensors such as EMG, which seems not exploited by the current version.

\section*{Conclusion}
This work introduced an automated sleep staging model called RobustSleepNet, designed for real-life use. Its architecture is close to previous deep learning methods such as SeqSleepNet and SimpleSleepNet; however, it differs from these by using attention layers to handle any arbitrary input shape. Thus, it can be used out of the box on any PSG montage. Using this property, we were able to train and evaluate the model in a leave one out dataset fashion on a corpus of 8 datasets. The model achieved a direct transfer performance close to the supervised one on most datasets. This performance is beyond any previous literature attempts; hence, it is likely to provide good results as is on unseen data. We highlight the need to train on populations with variability in age, montage and pathology to get the best performance on unseen data. An important parameter is using sleep scoring from multiple sleep experts to reduce the noise due to inter-rater variability.
We also demonstrated that using finetuning improves performances compared to learning from scratch — especially on the smallest datasets. One could use it to target specific populations. Further improvements could be made to RobustSleepNet to handle the specificity of the targeted record's montage. Finally, one could extend the model to providing metrics on the estimated confidence of the automated sleep staging.

% or


\section*{Acknowledgment}

\bibliographystyle{IEEEtran}
\bibliography{IEEEabrv,references}

\begin{thebibliography}{10}
\providecommand{\url}[1]{#1}
\csname url@samestyle\endcsname
\providecommand{\newblock}{\relax}
\providecommand{\bibinfo}[2]{#2}
\providecommand{\BIBentrySTDinterwordspacing}{\spaceskip=0pt\relax}
\providecommand{\BIBentryALTinterwordstretchfactor}{4}
\providecommand{\BIBentryALTinterwordspacing}{\spaceskip=\fontdimen2\font plus
\BIBentryALTinterwordstretchfactor\fontdimen3\font minus
  \fontdimen4\font\relax}
\providecommand{\BIBforeignlanguage}[2]{{%
\expandafter\ifx\csname l@#1\endcsname\relax
\typeout{** WARNING: IEEEtran.bst: No hyphenation pattern has been}%
\typeout{** loaded for the language `#1'. Using the pattern for}%
\typeout{** the default language instead.}%
\else
\language=\csname l@#1\endcsname
\fi
#2}}
\providecommand{\BIBdecl}{\relax}
\BIBdecl

\bibitem{Iber2007TheSpecifications}
C.~Iber, S.~Ancoli-Israel, A.~L. Chesson, and S.~F. Quan, \emph{{The AASM
  manual for the scoring of sleep and associated events : rules, terminology,
  and technical specifications}}.\hskip 1em plus 0.5em minus 0.4em\relax
  Westchester, IL: American Academy of Sleep Medicine, 2007.

\bibitem{Stephansen2018}
J.~Stephansen, A.~Olesen, M.~Olsen, A.~Ambati, E.~Leary, H.~Moore, O.~Carrillo,
  L.~Lin, F.~Han, H.~Yan, Y.~Sun, Y.~Dauvilliers, S.~Scholz, L.~Barateau,
  B.~Högl, A.~Stefani, S.~Hong, T.~Kim, F.~Pizza, and E.~Mignot, ``Neural
  network analysis of sleep stages enables efficient diagnosis of narcolepsy,''
  \emph{Nature Communications}, vol.~9, 12 2018.

\bibitem{Thorey_2019}
\BIBentryALTinterwordspacing
V.~Thorey, A.~B. Hernandez, P.~J. Arnal, and E.~H. During, ``Ai vs humans for
  the diagnosis of sleep apnea,'' \emph{2019 41st Annual International
  Conference of the IEEE Engineering in Medicine and Biology Society (EMBC)},
  Jul 2019. [Online]. Available:
  \url{http://dx.doi.org/10.1109/EMBC.2019.8856877}
\BIBentrySTDinterwordspacing

\bibitem{Rosenberg2014TheEvents}
R.~S. Rosenberg and S.~Van~Hout, ``{The American Academy of Sleep Medicine
  inter-scorer reliability program: Respiratory events},'' \emph{Journal of
  Clinical Sleep Medicine}, 2014.

\bibitem{guillot2019dreem}
A.~Guillot, F.~Sauvet, E.~H. During, and V.~Thorey, ``Dreem open datasets:
  Multi-scored sleep datasets to compare human and automated sleep staging,''
  2019.

\bibitem{10.1093/sleep/zsaa097}
\BIBentryALTinterwordspacing
P.~J. Arnal, V.~Thorey, E.~Debellemaniere, M.~E. Ballard, A.~Bou~Hernandez,
  A.~Guillot, H.~Jourde, M.~Harris, M.~Guillard, P.~Van~Beers, M.~Chennaoui,
  and F.~Sauvet, ``{The Dreem Headband compared to polysomnography for
  electroencephalographic signal acquisition and sleep staging},''
  \emph{Sleep}, 05 2020, zsaa097. [Online]. Available:
  \url{https://doi.org/10.1093/sleep/zsaa097}
\BIBentrySTDinterwordspacing

\bibitem{supratak2017deepsleepnet}
A.~Supratak, H.~Dong, C.~Wu, and Y.~Guo, ``Deepsleepnet: a model for automatic
  sleep stage scoring based on raw single-channel eeg,'' \emph{IEEE
  Transactions on Neural Systems and Rehabilitation Engineering}, vol.~25,
  no.~11, pp. 1998--2008, 2017.

\bibitem{SeqSleepNet}
\BIBentryALTinterwordspacing
H.~Phan, F.~Andreotti, N.~Cooray, O.~Y. Chen, and M.~De~Vos, ``Seqsleepnet:
  End-to-end hierarchical recurrent neural network for sequence-to-sequence
  automatic sleep staging,'' \emph{IEEE Transactions on Neural Systems and
  Rehabilitation Engineering}, vol.~27, no.~3, p. 400–410, Mar 2019.
  [Online]. Available: \url{http://dx.doi.org/10.1109/TNSRE.2019.2896659}
\BIBentrySTDinterwordspacing

\bibitem{perslev2019utime}
M.~Perslev, M.~H. Jensen, S.~Darkner, P.~J. Jennum, and C.~Igel, ``U-time: A
  fully convolutional network for time series segmentation applied to sleep
  staging,'' 2019.

\bibitem{phan2021xsleepnet}
H.~Phan, O.~Y. Chén, M.~C. Tran, P.~Koch, A.~Mertins, and M.~D. Vos,
  ``Xsleepnet: Multi-view sequential model for automatic sleep staging,'' 2021.

\bibitem{Optimal_transport_chambon}
\BIBentryALTinterwordspacing
S.~Chambon, M.~N. Galtier, and A.~Gramfort, ``{Domain adaptation with optimal
  transport improves EEG sleep stage classifiers},'' in \emph{{Pattern
  Recognition in Neuroimaging}}, Singapour, Singapore, Jun. 2018. [Online].
  Available: \url{https://hal.archives-ouvertes.fr/hal-01814190}
\BIBentrySTDinterwordspacing

\bibitem{MultipleChannelTrasnferLearning}
H.~Phan, O.~Y. Ch{\'e}n, P.~Koch, Z.~Lu, I.~V. McLoughlin, A.~Mertins, and
  M.~D. Vos, ``Towards more accurate automatic sleep staging via deep transfer
  learning,'' \emph{ArXiv}, vol. abs/1907.13177, 2019.

\bibitem{MASS}
C.~O'Reilly, N.~Gosselin, J.~Carrier, and T.~Nielsen, ``Montreal archive of
  sleep studies: an open-access resource for instrument benchmarking and
  exploratory research,'' \emph{Journal of sleep research}, vol.~23, 06 2014.

\bibitem{SleepEDF}
B.~{Kemp}, A.~H. {Zwinderman}, B.~{Tuk}, H.~A.~C. {Kamphuisen}, and J.~J.~L.
  {Oberye}, ``Analysis of a sleep-dependent neuronal feedback loop: the
  slow-wave microcontinuity of the eeg,'' \emph{IEEE Transactions on Biomedical
  Engineering}, vol.~47, no.~9, pp. 1185--1194, 2000.

\bibitem{SingleChannelMismatch}
\BIBentryALTinterwordspacing
H.~Phan, O.~Y. Ch{\'{e}}n, P.~Koch, A.~Mertins, and M.~D. Vos, ``Deep transfer
  learning for single-channel automatic sleep staging with channel mismatch,''
  \emph{CoRR}, vol. abs/1904.05945, 2019. [Online]. Available:
  \url{http://arxiv.org/abs/1904.05945}
\BIBentrySTDinterwordspacing

\bibitem{KLPersonalization}
\BIBentryALTinterwordspacing
H.~Phan, K.~Mikkelsen, O.~Y. Chén, P.~Koch, A.~Mertins, P.~Kidmose, and
  M.~De~Vos, ``Personalized automatic sleep staging with single-night data: a
  pilot study with kl-divergence regularization,'' \emph{Physiological
  Measurement}, May 2020. [Online]. Available:
  \url{http://dx.doi.org/10.1088/1361-6579/ab921e}
\BIBentrySTDinterwordspacing

\bibitem{bahdanau2014neural}
D.~Bahdanau, K.~Cho, and Y.~Bengio, ``Neural machine translation by jointly
  learning to align and translate,'' 2014.

\bibitem{pytorch}
A.~Paszke, S.~Gross \emph{et~al.}, ``Automatic differentiation in {PyTorch},''
  in \emph{NIPS Autodiff Workshop}, 2017.

\bibitem{GRU2014}
J.~Chung, {\c{C}}.~G{\"{u}}l{\c{c}}ehre, K.~Cho, and Y.~Bengio, ``Empirical
  evaluation of gated recurrent neural networks on sequence modeling,''
  \emph{CoRR}, vol. abs/1412.3555, 2014.

\bibitem{dropout}
\BIBentryALTinterwordspacing
N.~Srivastava, G.~Hinton, A.~Krizhevsky, I.~Sutskever, and R.~Salakhutdinov,
  ``Dropout: A simple way to prevent neural networks from overfitting,''
  \emph{Journal of Machine Learning Research}, vol.~15, no.~56, pp. 1929--1958,
  2014. [Online]. Available:
  \url{http://jmlr.org/papers/v15/srivastava14a.html}
\BIBentrySTDinterwordspacing

\bibitem{SEO2020102037}
\BIBentryALTinterwordspacing
H.~Seo, S.~Back, S.~Lee, D.~Park, T.~Kim, and K.~Lee, ``Intra- and inter-epoch
  temporal context network (iitnet) using sub-epoch features for automatic
  sleep scoring on raw single-channel eeg,'' \emph{Biomedical Signal Processing
  and Control}, vol.~61, p. 102037, 2020. [Online]. Available:
  \url{http://www.sciencedirect.com/science/article/pii/S1746809420301932}
\BIBentrySTDinterwordspacing

\bibitem{GraphSleepNet}
\BIBentryALTinterwordspacing
Z.~Jia, Y.~Lin, J.~Wang, R.~Zhou, X.~Ning, Y.~He, and Y.~Zhao, ``Graphsleepnet:
  Adaptive spatial-temporal graph convolutional networks for sleep stage
  classification,'' in \emph{Proceedings of the Twenty-Ninth International
  Joint Conference on Artificial Intelligence, {IJCAI-20}}, C.~Bessiere,
  Ed.\hskip 1em plus 0.5em minus 0.4em\relax International Joint Conferences on
  Artificial Intelligence Organization, 7 2020, pp. 1324--1330, main track.
  [Online]. Available: \url{https://doi.org/10.24963/ijcai.2020/184}
\BIBentrySTDinterwordspacing

\bibitem{varela}
I.~Fernández-Varela, E.~Hernández-Pereira, D.~Álvarez Estévez, and
  V.~Moret-Bonillo, ``A convolutional network for sleep stages
  classification,'' 02 2019.

\bibitem{Sors2018ACN}
A.~Sors, S.~Bonnet, S.~Mirek, L.~Vercueil, and J.-F. Payen, ``A convolutional
  neural network for sleep stage scoring from raw single-channel eeg,''
  \emph{Biomed. Signal Process. Control.}, vol.~42, pp. 107--114, 2018.

\bibitem{rechtschaffen1973manual}
\BIBentryALTinterwordspacing
A.~Rechtschaffen, A.~Kales, L.~A. B. I.~S. University~of California, and
  N.~N.~I. Network, \emph{A Manual of Standardized Terminology, Techniques and
  Scoring System for Sleep Stages of Human Subjects}, ser. Publication.\hskip
  1em plus 0.5em minus 0.4em\relax Brain Information Service/Brain Research
  Institute, University of California, 1973. [Online]. Available:
  \url{https://books.google.fr/books?id=cUvatQEACAAJ}
\BIBentrySTDinterwordspacing

\bibitem{sleep_edf_processing}
S.~A. {Imtiaz} and E.~{Rodriguez-Villegas}, ``An open-source toolbox for
  standardized use of physionet sleep edf expanded database,'' in \emph{2015
  37th Annual International Conference of the IEEE Engineering in Medicine and
  Biology Society (EMBC)}, 2015, pp. 6014--6017.

\bibitem{MrOS1}
J.~Blank, P.~Cawthon, M.~Carrion-Petersen, L.~Harper, J.~Johnson, E.~Mitson,
  and R.~Delay, ``Overview of recruitment for the osteoporotic fractures in men
  study (mros),'' \emph{Contemporary clinical trials}, vol.~26, pp. 557--68, 10
  2005.

\bibitem{MrOS2}
\BIBentryALTinterwordspacing
E.~Orwoll, J.~B. Blank, E.~Barrett-Connor, J.~Cauley, S.~Cummings, K.~Ensrud,
  C.~Lewis, P.~M. Cawthon, R.~Marcus, L.~M. Marshall, J.~McGowan, K.~Phipps,
  S.~Sherman, M.~L. Stefanick, and K.~Stone, ``Design and baseline
  characteristics of the osteoporotic fractures in men (mros) study — a large
  observational study of the determinants of fracture in older men,''
  \emph{Contemporary Clinical Trials}, vol.~26, no.~5, pp. 569 -- 585, 2005.
  [Online]. Available:
  \url{http://www.sciencedirect.com/science/article/pii/S1551714405001072}
\BIBentrySTDinterwordspacing

\bibitem{MrOS3}
\BIBentryALTinterwordspacing
T.~Blackwell, K.~Yaffe, S.~Ancoli-Israel, S.~Redline, K.~E. Ensrud, M.~L.
  Stefanick, A.~Laffan, K.~L. Stone, and for the Osteoporotic Fractures~in Men
  Study~Group, ``Associations between sleep architecture and sleep-disordered
  breathing and cognition in older community-dwelling men: The osteoporotic
  fractures in men sleep study,'' \emph{Journal of the American Geriatrics
  Society}, vol.~59, no.~12, pp. 2217--2225, 2011. [Online]. Available:
  \url{https://onlinelibrary.wiley.com/doi/abs/10.1111/j.1532-5415.2011.03731.x}
\BIBentrySTDinterwordspacing

\bibitem{NSSR1}
\BIBentryALTinterwordspacing
I.~Dean, Dennis~A., A.~L. Goldberger, R.~Mueller, M.~Kim, M.~Rueschman,
  D.~Mobley, S.~S. Sahoo, C.~P. Jayapandian, L.~Cui, M.~G. Morrical,
  S.~Surovec, G.-Q. Zhang, and S.~Redline, ``{Scaling Up Scientific Discovery
  in Sleep Medicine: The National Sleep Research Resource},'' \emph{Sleep},
  vol.~39, no.~5, pp. 1151--1164, 05 2016. [Online]. Available:
  \url{https://doi.org/10.5665/sleep.5774}
\BIBentrySTDinterwordspacing

\bibitem{NSRR2}
\BIBentryALTinterwordspacing
G.-Q. Zhang, L.~Cui, R.~Mueller, S.~Tao, M.~Kim, M.~Rueschman, S.~Mariani,
  D.~Mobley, and S.~Redline, ``The national sleep research resource: Towards a
  sleep data commons,'' in \emph{Proceedings of the 2018 ACM International
  Conference on Bioinformatics, Computational Biology, and Health Informatics},
  ser. BCB '18.\hskip 1em plus 0.5em minus 0.4em\relax New York, NY, USA:
  Association for Computing Machinery, 2018, p. 572. [Online]. Available:
  \url{https://doi.org/10.1145/3233547.3233725}
\BIBentrySTDinterwordspacing

\bibitem{SHHS1}
\BIBentryALTinterwordspacing
S.~F. Quan, B.~V. Howard, C.~Iber, J.~P. Kiley, F.~J. Nieto, G.~T. O'Connor,
  D.~M. Rapoport, S.~Redline, J.~Robbins, J.~M. Samet, and â.~W. Wahl, ``{The
  Sleep Heart Health Study: Design, Rationale, and Methods},'' \emph{Sleep},
  vol.~20, no.~12, pp. 1077--1085, 12 1997. [Online]. Available:
  \url{https://doi.org/10.1093/sleep/20.12.1077}
\BIBentrySTDinterwordspacing

\bibitem{SHHS2}
S.~Redline, M.~Sanders, B.~Lind, S.~Quan, C.~Iber, D.~Gottlieb, W.~Bonekat,
  D.~Rapoport, P.~Smith, and J.~Kiley, ``\BIBforeignlanguage{English
  (US)}{Methods for obtaining and analyzing unattended polysomnography data for
  a multicenter study},'' \emph{\BIBforeignlanguage{English (US)}{Sleep}},
  vol.~21, no.~7, pp. 759--767, Nov. 1998.

\bibitem{MultiChannelTransfer}
F.~{Andreotti}, H.~{Phan}, N.~{Cooray}, C.~{Lo}, M.~T.~M. {Hu}, and M.~{De
  Vos}, ``Multichannel sleep stage classification and transfer learning using
  convolutional neural networks,'' in \emph{2018 40th Annual International
  Conference of the IEEE Engineering in Medicine and Biology Society (EMBC)},
  July 2018, pp. 171--174.

\end{thebibliography}

\ifCLASSOPTIONcaptionsoff
  \newpage
\fi
\clearpage
\pagebreak
\appendices



%\bibliographystyle{IEEEtran}

\end{document}